\documentclass[conference]{IEEEtran}
\IEEEoverridecommandlockouts
\usepackage{cite}
\usepackage{amsmath,amssymb,amsfonts}
\usepackage{algorithmic}
\usepackage{graphicx}
\usepackage{textcomp}
\usepackage{xcolor}
\usepackage{hyperref}
\usepackage{amsmath, amssymb, algorithm, algorithmic}
\def\BibTeX{{\rm B\kern-.05em{\sc i\kern-.025em b}\kern-.08em
    T\kern-.1667em\lower.7ex\hbox{E}\kern-.125emX}}
\begin{document}

\title{ASD Classification on Dynamic Brain Connectome using Temporal Random Walk with Transformer-based Dynamic Network Embedding}
\author{Suchanuch Piriyasatit, Chaohao Yuan, Ercan Engin Kuruoglu$^*$\thanks{$^*$Corresponding author}\textit{ Senior Member, IEEE}
\thanks{Suchanuch Piriyasatit, Chaohao Yuan and Ercan Engin Kuruoglu are with Institute of Data and Information, Tsinghua Shenzhen International Graduate School, Shenzhen, Guangdong, China (z/-yt22@mails.tsinghua.edu.cn; yuanch22@mails.tsinghua.edu.cn; kuruoglu@sz.tsinghua.edu.cn).}
\thanks{This work is supported by Shenzhen Science and Technology Innovation Commission under Grant JCYJ20220530143002005, Tsinghua University SIGS Start-up fund under Grant QD2022024C, and Shenzhen Ubiquitous Data Enabling Key Lab under Grant ZDSYS20220527171406015.}

}

\maketitle

\begin{abstract}
Autism Spectrum Disorder (ASD) is a complex neurological condition characterized by varied developmental impairments, especially in communication and social interaction. Accurate and early diagnosis of ASD is crucial for effective intervention, which is enhanced by richer representations of brain activity. The brain functional connectome, which refers to the statistical relationships between different brain regions measured through neuroimaging, provides crucial insights into brain function. Traditional static methods often fail to capture the dynamic nature of brain activity, in contrast, dynamic brain connectome analysis provides a more comprehensive view by capturing the temporal variations in the brain. We propose BrainTWT, a novel dynamic network embedding approach that captures temporal evolution of the brain connectivity over time and considers also the dynamics between different temporal network snapshots. BrainTWT employs temporal random walks to capture dynamics across different temporal network snapshots and leverages the Transformer's ability to model long term dependencies in sequential data to learn the discriminative embeddings from these temporal sequences using temporal structure prediction tasks. The experimental evaluation, utilizing the Autism Brain Imaging Data Exchange (ABIDE) dataset, demonstrates that BrainTWT outperforms baseline methods in ASD classification.
\end{abstract}


\section{Introduction}
Autism Spectrum Disorder (ASD) is a developmental disorder characterized by difficulties in social interaction, repetitive behaviors, and restricted interests, with symptoms varying widely across the spectrum. The classification of ASD based on brain characteristics has advanced significantly with the evolution of neuroimaging technologies and analytical techniques. Historically, the diagnosis of ASD relied heavily on behavioral assessments and clinical observations \cite{mirenda2013autism, mintz2017evolution}. However, with the introduction of imaging techniques like MRI and fMRI, researchers have been able to study brain structure and function in more detail \cite{liu2021autism, patil2024neuroimaging, giansanti10umbrella}.

Brain functional connectivity or functional connectome refers to statistical relationships between different regions of the brain measured by correlation of neural signals. This connectivity can be observed through various neuroimaging techniques like fMRI, which detect blood flow changes associated with neural activity, indicating how different parts of the brain communicate during tasks or in resting states. Functional connectivity provides insights into both normal brain function and the impacts of neurological disorders on the brain. Brain's functional states are not static but evolve over time \cite{lee2024human}, hence there is the need to study dynamic functional connectivity to understand the complex brain activity related to disorders like ASD. Studies \cite{harlalka2019atypical, ma2022dynamic, chentda2024} have highlighted how dynamic functional connectivity provides deeper insights into the temporal variability of brain networks, revealing patterns that static connectivity measures might miss. 

Given a dynamic functional connectivity of the brain, the task of detecting ASD can be approached as a dynamic network classification task, which allows for the use of network embedding techniques. Network embedding techniques have become essential in capturing the complexities of network structures by transforming a network into a low-dimensional space for downstream tasks. A node-level embedding transforms each node in a network to a vector representation for node-level tasks such as node classification, link prediction, and node clustering while a graph-level or network-level embedding methods transform the whole network into a single vector representations that summarizes the entire network structure for graph-level tasks such as graph classification, graph similarity ranking, and anomaly detection.

Static network embedding methods often fail to capture the dynamic nature of the brain connectome. On the other hand, majority of the graph-level dynamic network embedding methods model the dynamic network in a snapshot-based manner where an embedding for each temporal network snapshot is learned independently without consideration for interactions or dynamics between snapshots. To address this, we propose \textbf{BrainTWT}, a novel graph-level dynamic network embedding that, instead of considering the dynamic brain network at each time snapshot independently, captures the dynamics across different time snapshots and models temporal nodes' interactions attention on temporal structure prediction tasks. This is achieved by leveraging temporal random walks to capture the inter-snapshot dynamics of the dynamic brain network and utilizing a Transformer-based model to model dependencies and interactions across network states over time. The experimental results using the Autism Brain Imaging Data Exchange (ABIDE) benchmarking dataset \cite{di2014autism} show that BrainTWT achieves better performance compared to other baselines for ASD classification.

\section{Related Work}
Analysis on fMRI signals and functional connectivity data for brain disorder classification including ASD has advanced with the development of machine learning techniques. Initial studies focused on classical models such as Support Vector Machines (SVM) and Random Forest to analyze correlations between different brain regions evident in functional MRI (fMRI) data \cite{abraham2017deriving, subbaraju2017identifying}. Given high-dimensional nature of fMRI data, deep learning models frameworks are later applied including Convolutional Neural Networks, Recurrent Networks and have been employed to explore both spatial and temporal aspects of brain functional connectivity \cite{khosla20183d, thomas2020classifying,ahammed2021darkasdnet}.

Another branch of approaches applied to functional connectivity is through using network embedding methods. Network embedding methods can be divided into node-level and network-level (or graph-level) embeddings, each can be designed for static or dynamic network:\\
\textbf{Node-level Embedding}: Node-level embedding methods learn a vector representation of each node in a network for node-level tasks such as node classification and link prediction. Various methods have been proposed for static networks. Matrix factorization-based method such as Locally Linear Embedding (LLE) \cite{roweis2000nonlinear}, Laplacian eigenmaps \cite{belkin2001laplacian}, SVD \cite{golub1971singular}, HOPE \cite{ou2016asymmetric}, and GraRep \cite{cao2015grarep}, decompose the Laplacian or network adjacency matrix to create embeddings that preserve node structures and high-order proximities. Random walk-based methods such as DeepWalk \cite{perozzi2014deepwalk} and node2vec \cite{grover2016node2vec} use random walks to sample the context of nodes and apply Skip-gram to learn the embeddings. Moreover, Graph neural networks (GNNs) including graph convolutional networks (GCN) \cite{kipf2016semi}, graph attention networks (GAT) \cite{velickovic2017graph}, graph auto-encoders (GAEs) \cite{kipf2016variational} leverage message-passing mechanisms to generate network embeddings. To learn embeddings for dynamic network, various methods build upon these static methods to effectively learn node embeddings in time-varying networks. This includes CTDNE \cite{nguyen2018dynamic} which extends DeepWalk by using temporal random walks with Skip-gram. Dynamic graph neural networks are proposed such as EvolveGCN \cite{pareja2020evolvegcn}, TGAT \cite{xu2020inductive}.\\ 
\textbf{Network-level or Graph-level embedding}: Graph-level embedding methods learn a vector representation of the entire network for graph-level task such as graph classification. Various approaches use the results from node-level embeddings and aggregate all the node embeddings, by maximizing, averaging, or summing, into one vector to represent the whole network \cite{li2024graph}. While other methods are proposed to directly learn embedding for the whole network. Early graph kernels methods such as message passing kernels \cite{shervashidze2011weisfeiler, neumann2016propagation}, shortest path kernels \cite{borgwardt2005shortest, nikolentzos2017shortest}, and subgraph kernels \cite{shervashidze2009efficient, kriege2012subgraph}, focus on capturing graph isomorphism and substructure frequencies. Skip-gram-based and random walk-based methods \cite{narayanan2016subgraph2vec, adhikari2018sub2vec} use sequences of nodes with Skip-gram to learn embeddings. Deep learning methods include recurrent neural network-based methods (RNNs) \cite{zhao2018substructure}, convolution neural network-based methods (CNNs) \cite{simonovsky2017dynamic, nikolentzos2018kernel}, GNNs-based method \cite{morris2019weisfeiler,nikolentzos2020random}. Recently the Transformer-based methods \cite{yuan2025survey} have gained increasing popularity. More in-depth reviews and analysis can be found in \cite{yang2024state}. On the other hand, for dynamic networks, graph-level embedding methods without having to resort to pooling of node embeddings, have not been as widely explored. tdGraphEmbed \cite{beladev2020tdgraphembed} is the first to propose graph-level method based on random walk with Skip-gram to learn an embedding of each time snapshot of a dynamic network. GraphERT \cite{beladev2023graphert} uses random walk sequences sampled independently from each temporal network snapshot and utilizes a Transformer-based model to learn the embeddings. However, both tdGraphEmbed and GraphERT sample random walks independently from each timestep with the goal to obtain network representation for each temporal network snapshot. Their downstream tasks are conducted within different time snapshots of a dynamic network (e.g. similarity ranking, anomaly detection). Although we have different goals of summarizing the entire dynamic network and considering inter-snapshot relationships within a dynamic network, we are motivated by GraphERT to use the Transformers to model the inter-snapshot relationships where the learned embedding will have information of the entire dynamic brain connectivity across different time steps.\\
% Motivated by GraphERT, we apply temporal random walk to sample dynamic sequences of the entire dynamic network and leverage the ability to model long-term dependencies of Transformers to learn the embedding of the entire dynamic network for ASD classification\\
\textbf{Applications of network embedding for brain networks and ASD classification}: \cite{yousefian2023detection} explores various static network embedding methods applied to fMRI data for classifying ASD in a static functional connectivity. The methods explored are random-walk based methods including node2vec \cite{grover2016node2vec}, struc2vec \cite{ribeiro2017struc2vec}, AWE \cite{ivanov2018anonymous}.  t-BNE \cite{cao2017t} proposes tensor factorization-based method employing a partially symmetric tensor factorization approach with side information guidance to capture meaningful patterns associated with brain disorders. Various methods leverage graph neural networks to learn the brain functional connectivity. Hi-GCN \cite{jiang2020hi} learns the brain network embeddings using hierarchical GCN to capture topological patterns in individual brain network considering relations with broader population-level characteristics. Other approaches are proposed to consider dynamic brain networks. \cite{gadgil2020spatio} explores the age and sex difference on resting-state fMRIs using Spatio-temporal graph convolution network (ST-GCN) \cite{yan2018spatial, yu2017spatio}. Sequential Monte Carlo GCN (SMC-GCN) \cite{zhao2024sequential} adopts particle filtering and allows for a statistical interpretation of dynamic functional connectivity. \cite{noman2022graph} proposes Graph Autoencoder-based embedding to learn dynamic brain networks for ASD classification.
% \cite{lin2021learning} combines harmonic waves and Fourier bases to represent the topology and temporal dynamics of functional connectivities.  
% %
\section{Problem Statement}
Given a set of dynamic brain connectomes \(\mathcal{G} = \{G_1, G_2, \ldots, G_M\}\), where \(G_i = (V_i, E_i)\) represents a dynamic brain connectome of subject $i$, and $V_i$ is the set of nodes shared across time steps and $E_i \subseteq V_i \times V_i \times T_i$ is the set of all temporal edges in $G_i$. Our goal is to find a mapping function $f : G \rightarrow \mathbb{R}^d$ that transform each dynamic connectome $G_i$ into a $d$-dimensional vector space. The resulting embeddings will be used as inputs to train logistic classifier for the ASD classification.

\section{Methodology}

\begin{figure*}
\begin{center}
 \includegraphics[width=\textwidth]{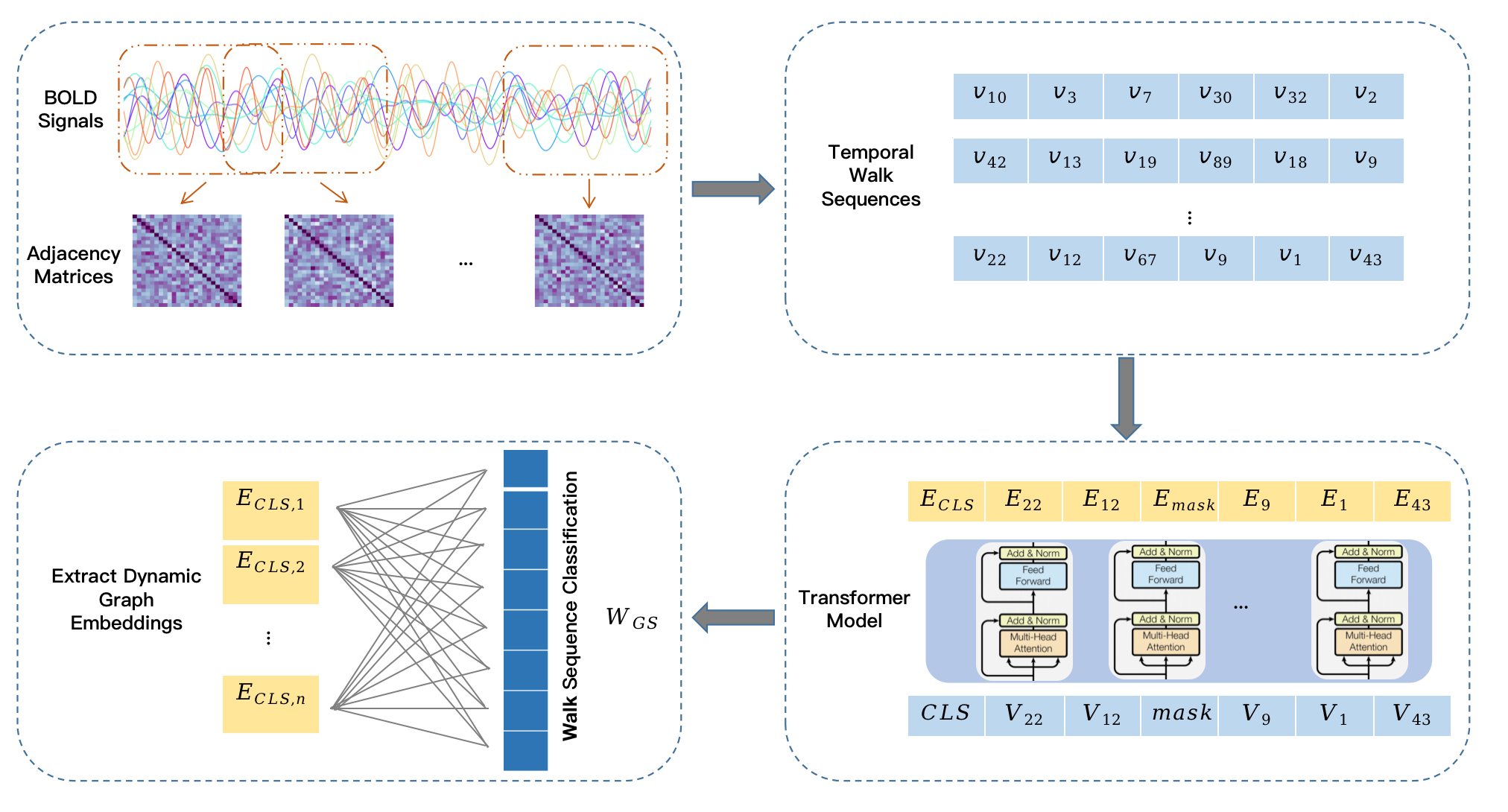}
 \caption{The dynamic brain network is obtained from Pearson’s correlation between the BOLD time-series of each pair of regions of interest (ROIs) in the brain. The dynamic brain network is then converted into a series of temporal random walk sequences, capturing the temporal evolution of brain connectivity. Each temporal sequence is then tokenized and embedded. The embedded sequences are processed through a Transformer model that uses self-attention mechanisms on node interactions based on their temporal and contextual significance. Parts of the sequence are masked, and the model predicts these masked nodes, refining the embeddings using contextual data. The model incorporates joint learning to optimize both the temporal dynamics and graph-level loss. The final embeddings are used as input features for ASD classification task.}
\end{center}
\end{figure*}

\subsection{Sampling Temporal Sequences with Temporal Random Walk}
We first convert a dynamic brain network $G_i$ of subject $i$ into a set of temporal random walk sequences $\{\mathcal{W}_{1}, \mathcal{W}_{2}, ... \mathcal{W}_{n}\}$ where $n$ is the number of random walk samples per network. Note that we drop the network index in the random walk sequence notation for simplicity.

A temporal random walk modifies the traditional random walk and incorporates time into the walk to ensure the transitions follow the chronological sequence of events that occur in the dynamic network. Formally, a temporal walk $\mathcal{W}_k$ on a dynamic network $G_i$ is a sequence of nodes $\{v_{1}, v_{2}, ..., v_{l}\}$ of length $l$, where the edge $e_i$, which is the edge connecting $v_{ i}$ and $v_{i+1}$ for $ 1 \leq i < l-1$, satisfies $\mathcal{T}(e_i) \leq \mathcal{T}(e_{i+1})$ where $\mathcal{T}(e_i)$ and $\mathcal{T}(e_{i+1}) \in 1 \dots T$ are the timesteps of edges $e_i$ and $e_{i+1}$, respectively. This incorporates time information into the walk and ensures that the walk is a forward move in time and that it is an ordered sequence of events in the network.

Consider a random walker positioned at node \(v_j\) during time \(t\). The set of temporal edges connected to node \(v_j\) at time \(t\) can be defined as:
\[
\mathcal{N}_T(v_j, t) = \{ e_{jk}^{t'} = (v_j, v_k, t') \in E_i  \text{ where } t' \geq t \}.
\]
The probability of transitioning along an edge within this neighborhood is given by:
\begin{equation}
 \mathcal{P}_T(e_{jk}^{t'}|v_j, t) = \frac{\exp[t-t']}{\sum_{e \in \mathcal{N}_T(v_j, t)} \exp[t-\mathcal{T}(e)]}.
\end{equation}
This choice of the exponential function prioritizes edges that are closer in time to the current position, thereby reducing the likelihood of the walker making temporally distant jumps and preserving the continuity of time in the walk.

\subsection{Learn Temporal Dynamics with the Transformer Model}
To learn temporal dynamics from sequences of temporal random walk, we mask a percentage of nodes in each temporal sequence, and then predict those masked nodes, where we use the Transformer to model the token embeddings. This masked sequence prediction has been done in the context of natural language processing \cite{kenton2019bert} which we will follow.

 Before going through the self-attention layers, an input temporal walk sequence $\mathcal{W}_k = \{v_{k,1}, v_{k,2}, ..., v_{k,l}\} \in \mathbb{R}^l$ will be tokenized and initialized as an input embedding $X_k = (<CLS>, \textbf{v}_{k,1}, \textbf{v}_{k,2}, ..., \textbf{v}_{k,l}) \in$ $\mathbb{R}^{n \times d}$, where $n = l+1$ is the sequence length with the added special tokens $<CLS>$ to specify the start of a sequence. What important about the $<CLS>$ token is that its final hidden state after going through all the attention layers can be used as the aggregate embedding of this temporal sequence \cite{kenton2019bert}. Next, we state the rest of the Transformer model \cite{vaswani2017attention}:

\subsubsection{Self-Attention Mechanism}
The self-attention mechanism dynamically determines the relevance, or ``attention," that each token in the sequence should pay to every other token when constructing its own representation. Given an input sequence embedding $X_k = (<CLS>, \textbf{v}_{k,1}, \textbf{v}_{k,2}, ..., \textbf{v}_{k,l}) \in$ $\mathbb{R}^{n \times d}$, the attention scores is then computed using three learnable weight matrices \(W^Q\), \(W^K\), \(W^V \in \mathbb{R}^{d\times d_k}\) as
\begin{equation}
Q = XW^Q, \quad K = XW^K, \quad V = XW^V. \label{kqv}
\end{equation}
The attention function then allows each node to dynamically attend to every node in the walk sequence, weighted by their computed attention scores,
\begin{equation}
\text{Attention}(Q, K, V) = \text{softmax}\left(\frac{QK^T}{\sqrt{d_k}}\right)V. \label{self_attention}
\end{equation}
This updates the node embedding to reflect not only its current connections but also other connections along the temporal walk over time. Moreover, the Transformer can consider multiple sets of these learnable weight matrices, i.e.,  \(W_i^Q\), \(W_i^K\), \(W_i^V \in \mathbb{R}^{d\times d_k}\),  and concatenate their attention outputs together, as called multi-heads attention, to capture multiple patterns in the data:
\begin{equation}
\text{MultiHead}(Q, K, V) = \text{Concat}(\text{head}_1, ..., \text{head}_h)W^O, \label{multihead}
\end{equation}
\noindent \text{where} $W^O \in \mathbb{R}^{hd_k \times d}$, $h$ is the attention heads considered, and
\begin{equation}
\text{head}_i = \text{Attention}(QW_i^Q, KW_i^K, VW_i^V). \label{head_i}
\end{equation}
\subsubsection{Position-wise Feed-Forward Networks}
Following the attention mechanism, each position's output is independently fed through a position-wise feed-forward network, which is essentially two linear transformations with a ReLU activation in between:
\begin{equation}
\text{FFN}(x) = \max(0, xW_1 + b_1)W_2 + b_2, \label{feedforward}
\end{equation}
\noindent \text{where} $W_1 \in \mathbb{R}^{d \times d_{ff}}$, $b_1 \in \mathbb{R}^{d_{ff}}$, $W_2 \in \mathbb{R}^{d_{ff} \times d}$, $b_2 \in \mathbb{R}^{d}$.
\subsubsection{Position Encodings}
To account for the sequence order, positional encodings are added to the input embeddings. We use the positional encoding proposed in the original Transformer paper \cite{vaswani2017attention}:
\begin{equation}
\text{PE}(pos, 2i) = \sin\left(\frac{pos}{10000^{2i/d}}\right) \label{pos_encoding_sin}
\end{equation}
\begin{equation}
\text{PE}(pos, 2i+1) = \cos\left(\frac{pos}{10000^{2i/d}}\right), \label{pos_encoding_cos}
\end{equation}
where $pos$ is the token position and $i$ is the index $ \in 1...d$ of the embedding dimension.

\begin{table*}[ht]
\caption{10-fold Cross Validation Performance Comparison}
\centering
\footnotesize % You can also try \scriptsize or \footnotesize for even smaller text\cite{cao2022modeling}
\begin{tabular}{c c c c c}
\hline
 \noalign{\smallskip}
\textbf{Method} & \textbf{Accuracy} & \textbf{Sensitivity} & \textbf{Specificity} & \textbf{AUC} \\[0.25ex]
 \noalign{\smallskip}
\hline
 \noalign{\smallskip}
Random Forest \cite{breiman2001random} &   0.6038 $\pm$  0.0572     &     0.3926 $\pm$ 0.0751   &  \textbf{  0.7860 $\pm$ 0.0547 }     &    0.6660 $\pm$ 0.0512    \\  \noalign{\smallskip}
GCN  \cite{kipf2016semi}   &   0.6246  $\pm$  0.0593    &   0.5753 $\pm$ 0.1463     &   0.6659 $\pm$ 0.1498     &   0.6445 $\pm$ 0.0589  \\ \noalign{\smallskip}
GAT  \cite{velickovic2017graph}   &   0.6302 $\pm$ 0.0433   &     0.5527 $\pm$ 0.1285   &  0.6960 $\pm$ 0.1217    &  0.6551 $\pm$ 0.0651 \\ \noalign{\smallskip}
tdGraphEmbed  \cite{beladev2020tdgraphembed}    &   0.6325 $\pm$ 0.0432  &   0.5656 $\pm$ 0.0534    &  0.6902 $\pm$ 0.0509   & 0.6671 $\pm$ 0.0437   \\ \noalign{\smallskip}
GraphERT  \cite{beladev2023graphert}    &  0.6338 $\pm$ 0.0694  &  0.4648 $\pm$ 0.0937     &  0.7794 $\pm$ 0.0945  & 0.6949 $\pm$ 0.0712  \\  \noalign{\smallskip}
GSA-LSTM   \cite{cao2022modeling}     &   0.6840   &    0.6440     &    0.6980  &    0.7050    \\ \noalign{\smallskip}
\hline
\noalign{\smallskip}
BrainTWT \\ (without temporal dynamics)       &   0.5074 $\pm$ 0.0461  &   0.3453 $\pm$ 0.0821     &   0.6475 $\pm$ 0.0756     &    0.5093 $\pm$ 0.0664     \\ \noalign{\smallskip}
\textbf{BrainTWT} \\ \textbf{(with temporal dynamics)}       &  \textbf{0.7003  $\pm$ 0.0494}   &   \textbf{ 0.6504 $\pm$ 0.0733 }   &     0.7435 $\pm$ 0.0969   &    \textbf{0.7527 $\pm$ 0.0445 }   \\
 \noalign{\smallskip}
\hline
\end{tabular}
\label{tab:10-fold-result}
\end{table*}

\subsection{Temporal Dynamics and Graph-level Loss Joint Learning}

The joint loss in the similar manner as \cite{beladev2023graphert} is adopted. The main difference comes from the use of temporal random walk that makes a sequence includes nodes from across different timesteps of the network in temporal order. Therefore, instead of learning network structures through random walk path within each time snapshot independently, we learn temporal dynamics of the entire network considering its inter-snapshot relationships. Finally, we will obtain an embedding to represent the whole dynamic network.\\
\textbf{The Temporal Dynamics Loss}: A one-layer MLP classifier $C_{temporal}: \mathbb{R}^d \rightarrow \mathbb{R}^{|V|}$ with learnable weight $W_{TD} \in \mathbb{R}^{d \times |V|}$ is constructed to predict the masked token in the temporal walk sequences. Given a masked token's embedding output, $E_i$, from the Transformer, the classifier predicts the actual node that corresponds to the token. The temporal dynamics loss is then the cross-entropy loss of the classifier:
\begin{equation}
\mathcal{L}_{TD} = -\frac{1}{n} \sum_{i=1}^n \sum_{(j,k)} \log((E_{i,j} \cdot W_{TD})_k), \label{mlm_loss}
\end{equation}
where $n$ is the number of random walk samples, $E_{i,j}$ is the embedding output from the Transformer of the $j^{\text{th}}$ token in a walk sequence $i$, and $k$ is the index of the actual node that corresponds to the masked postition.\\
\textbf{The Graph-Level Loss}:
Intuitively, each learned embedding for each walk sequence should capture dynamics specific to its original temporal graph. Given an embedding of a random walk sequence, we want to predict the dynamic brain network from which it is sampled. Similarly, another one-layer MLP classifier $C_{graph}: \mathbb{R}^d \rightarrow \mathbb{R}^{|\mathcal{G}|}$ with learnable weight $W_{GS} \in \mathbb{R}^{d \times |\mathcal{G}|}$ is constructed to predict probability distribution of all dynamic brain networks a sequence is sampled from.

The $<CLS>$ token embedding output, $E_{CLS_i}$, from the Transformer for each random walk sequence is used as the aggregated representation for that walk sequence since it has aggregated contextual information from the entire sequence through the self-attention mechanism. The graph-level loss is
\begin{equation}
\mathcal{L}_{GS} = -\frac{1}{n} \sum_{i=1}^n \log((E_{CLS_i} \cdot W_{GS})_k) \label{eq:loss_function}
\end{equation}
where $k$ is the index of the true dynamic network.\\
\textbf{The Joint Loss} is then as follows:
\begin{equation}
\mathcal{L}_{\text{total}} = \lambda_1 \cdot \mathcal{L}_{TD} + \lambda_2 \cdot \mathcal{L}_{GS},
\label{eq:final_loss}
\end{equation}
where $\lambda_1$ and $\lambda_2$ are trade-offs between temporal dynamics and graph-level loss. This joint loss enables the model to learn temporal dynamics within each dynamic network, while ensuring that the learned embeddings are discriminative and uniquely represent each network.

After the training, $W_{GS}(G_i)$ is the resulting dynamic graph embedding of the dynamic graph $G_i$. These resulting embeddings are then used as inputs with corresponding ASD labels to train a logistic classifier.

\section{Experiment and Results}
Our code is publicly available at  \href{https://github.com/suchanuchp/BrainTWT}{https://github.com/suchanuchp/BrainTWT}.

\subsection{Data Preparation}
The resting state fMRI data is from the Autism Brain Imaging Data Exchange (ABIDE) \cite{di2014autism}. A total of 871 subjects are considered where the subjects are filtered by experts to include only high quality data subjects result in a total of 871 subjects with 468 healthy controls and 403 ASDs \cite{abraham2017deriving}. The data is preprocessed using standardized method \cite{nielsen2013multisite}, including slice timing correction, motion correction, motion scrubbing, etc. The BOLD time-series for each regions of interest (ROIs) is extracted using the AAL atlas \cite{tzourio2002automated} which partitions the brain into 116 ROIs. To get a dynamic brain network for each subject, Pearson’s correlation between the BOLD time-series of each pair of ROIs is calculated, thresholding method is applied to keep only the connections where the correlation exceeds the 80$^{th}$ percentile. This is done on a sliding window of length 50 and stride 5 to obtain a dynamic brain network.

\subsection{Experimental Settings}
We set the maximum length of random walk $l=20$ and sample 30 walks per node. We mask 15\% of the nodes for the temporal masked sequence prediction task. For the Transformer, we set the embedding dimension $d = 252$, the number of attention heads $h = 4$, and 6 hidden layers. To demonstrate the importance of temporal dynamics modeling, an ablation test is done by showing BrainTWT without the temporal dynamics loss ($\lambda_1=0$) alone with BrainTWT with the temporal dynamics loss ($\lambda_1=1$, $\lambda_2=5$).

\subsection{Evaluation and Baseline Methods}
We use the stratified K-fold cross-validation with $K=10$ as the evaluation method. Stratified K-fold cross-validation divides the dataset into $K$ folds and maintains an equal proportion of each class label in every fold to preserve the original distribution. Each fold serves once as a validation set while the model trains on the remaining folds. The mean scores from the cross-validation are reported along with the standard deviations. The results are compared with six baseline models including Random Forest \cite{breiman2001random}, GCN \cite{kipf2016semi}, GAT \cite{velickovic2017graph}, tdGraphEmbed \cite{beladev2020tdgraphembed}, GraphERT \cite{beladev2023graphert}, and GSA-LSTM \cite{cao2022modeling}.  We use mean pooling on node embeddings from GCN and GAT to get a network embedding. For tdGraphembed and GraphERT, we use mean pooling on temporal network snapshot embeddings to get an aggregated embedding of the entire dynamic network. Additionally, we also conduct a leave-one-site-out cross validation on our model for references. This is when all subjects from one site are used for evaluation and all subjects from the other sites are used for training.

\subsection{Results and Discussions}
The results of 10-fold stratified cross-validation are shown in Table~\ref{tab:10-fold-result}. BrainTWT, when incorporating temporal dynamics learning, achieves a relative increase of 2.38\%, 0.99\%, 6.52\%, and 6.77\%, in accuracy, sensitivity, specificity, and AUC, respectively, when compared to the second best performing model, GSA-LSTM. In ablation study, it also significantly outperforms its counterpart that excludes the temporal dynamics learning. This demonstrates the importance of the temporal dynamics loss to learn the network embedding. Furthermore, it outperforms the static baselines, Random Forest, GCN, and GAT, again showing the importance of brain temporal information in ASD classification. Lastly, it outperforms the snapshot-based dynamic methods, tdGraphEmbed and GraphERT, showing the importance of including the inter-snapshot dynamics into the embeddings. Results of the leave-one-site out validation for BrainTWT can be found for reference in Table~\ref{tab:leave-one-out} where the highest accuracy is from the site \textit{TRINITY} with 88.64\% accuracy and the lowest is from \textit{STANFORD} with 52.00\% accuracy

\begin{table}[h]
\caption{leave-one-site-out results}
\centering
\begin{tabular}{c c c c c c}
\hline
 \noalign{\smallskip}
\textbf{Site} & \textbf{Subject} & \textbf{Acc.} & \textbf{Sen.} & \textbf{Spe.} & \textbf{AUC} \\
              & \textbf{Count}   &               &               &               &              \\  \noalign{\smallskip}\hline
 \noalign{\smallskip}           
     CMU      & 11               & 0.8181        & 0.8333        & 0.8000        & 0.8000       \\ \noalign{\smallskip}
 CALTECH      & 15               & 0.8666        & 0.6000        & 1.0000        & 0.7200       \\ \noalign{\smallskip}
     KKI      & 33               & 0.7576        & 0.7500        & 0.7619        & 0.7500       \\ \noalign{\smallskip}
LEUVEN\_1     & 28               & 0.6786        & 0.5714        & 0.7857        & 0.7806       \\ \noalign{\smallskip}
LEUVEN\_2     & 28               & 0.6786        & 0.5833        & 0.7500        & 0.7448       \\ \noalign{\smallskip}
 MAX\_MUN     & 46               & 0.5870        & 0.5263        & 0.6296        & 0.5867       \\ \noalign{\smallskip}
     NYU      & 172              & 0.6453        & 0.8108        & 0.5204        & 0.7625       \\ \noalign{\smallskip}
    OHSU      & 25               & 0.5200        & 0.4167        & 0.6154        & 0.6154       \\ \noalign{\smallskip}
    OLIN      & 28               & 0.6429        & 0.7143        & 0.5714        & 0.7194       \\ \noalign{\smallskip}
    PITT      & 50               & 0.6800        & 0.4583        & 0.8846        & 0.7180       \\ \noalign{\smallskip}
     SBL      & 26               & 0.6538        & 0.5000        & 0.7857        & 0.6905       \\ \noalign{\smallskip}
    SDSU      & 27               & 0.7407        & 0.5000        & 0.8421        & 0.8355       \\ \noalign{\smallskip}
STANFORD      & 25               & 0.5200        & 0.5833        & 0.4615        & 0.6154       \\ \noalign{\smallskip}
 TRINITY      & 44               & 0.8864        & 0.9474        & 0.8400        & 0.9116       \\ \noalign{\smallskip}
  UCLA\_1     & 64               & 0.6406        & 0.6486        & 0.6296        & 0.6947       \\ \noalign{\smallskip}
  UCLA\_2     & 21               & 0.8571        & 1.0000        & 0.7000        & 0.8364       \\ \noalign{\smallskip}
    UM\_1     & 86               & 0.5814        & 0.5882        & 0.5769        & 0.6572       \\ \noalign{\smallskip}
    UM\_2     & 34               & 0.7647        & 0.5385        & 0.9048        & 0.7912       \\ \noalign{\smallskip}
     USM      & 67               & 0.6418        & 0.5581        & 0.7917        & 0.7965       \\ \noalign{\smallskip}
    YALE      & 41               & 0.7561        & 0.7273        & 0.7895        & 0.8062       \\ \noalign{\smallskip}
\hline
\end{tabular}
\label{tab:leave-one-out}
\end{table}

\section{Conclusion}
In this study, we propose a dynamic network embedding method BrainTWT for ASD classification that leverages temporal random walks and Transformer-based models to capture the dynamic evolution of brain connectivity over time. Our method outperforms traditional static and snapshot-based dynamic methods by effectively incorporating the temporal dynamics and inter-snapshot relationships within dynamic brain functional connectomes. The evaluation using the ABIDE dataset confirms that BrainTWT significantly improves classification performance, demonstrating the importance of modeling temporal information in neurological disorder analysis. Future research can extend these techniques to other cognitive disorders and integrate multimodal neuroimaging data to enrich the diagnostic potential. In the future, we will also exploit spectral graph information for the dynamic network prediction method \cite{yan2022adaptive}.

\bibliographystyle{plain}
\bibliography{ref.bib}

\end{document}